\documentclass[10pt,twocolumn,letterpaper]{article}

\usepackage{wacv}
\usepackage{times}
\usepackage{epsfig}
\usepackage{graphicx}
\usepackage{amsmath}
\usepackage{amssymb}
\usepackage{booktabs}
\usepackage{cite}

\usepackage{booktabs}
\usepackage{bbm}
\usepackage[table, dvipsnames]{xcolor}
\usepackage{caption}
\usepackage{subcaption}
\usepackage{adjustbox}
\usepackage[accsupp]{axessibility}

\def\ie{{\em i.e.}}
\def\etal{{\em et~al.}}

\newcommand\doubleplus{+\kern-1.3ex+\kern0.8ex}

\def\wacvPaperID{1033} % Enter the WACV Paper ID here

\wacvalgorithmstrack   % Uncomment this line if you are submitting to the Algorithms Track.
\wacvfinalcopy % *** Uncomment this line for the final submission

\ifwacvfinal
\usepackage[breaklinks=true,bookmarks=false,colorlinks=true]{hyperref}
\else
\usepackage[pagebackref=true,breaklinks=true,colorlinks=true,bookmarks=false]{hyperref}
\fi

\begin{document}

%%%%%%%%% TITLE
\title{Reconstructing Humpty Dumpty: \\Multi-feature Graph Autoencoder for Open~Set Action Recognition}

\author{Dawei Du~~~Ameya Shringi~~~Anthony Hoogs~~~Christopher Funk\\[3pt]
% \vspace{-5pt}\\
\large \textsl{Kitware}\\
{\tt\small \url{https://github.com/Kitware/graphautoencoder}}
}

\maketitle
\thispagestyle{empty}

\begin{abstract}
Most action recognition datasets and algorithms assume a closed world, where all test samples are instances of the known classes. In open~set problems, test samples may be drawn from either known or unknown classes. Existing open~set action recognition methods are typically based on extending closed~set methods by adding post hoc analysis of classification scores or feature distances and do not capture the relations among all the video clip elements. Our approach uses the reconstruction error to determine the novelty of the video since unknown classes are harder to put back together and thus have a higher reconstruction error than videos from known classes. We refer to our solution to the open~set action recognition problem as ``Humpty Dumpty'', due to its reconstruction abilities. Humpty Dumpty is a novel graph-based autoencoder that accounts for contextual and semantic relations among the clip pieces for improved reconstruction. A larger reconstruction error leads to an increased likelihood that the action can not be reconstructed, \ie, can not put Humpty Dumpty back together again, indicating that the action has never been seen before and is novel/unknown. Extensive experiments are performed on two publicly available action recognition datasets including HMDB-51 and UCF-101, showing the state-of-the-art performance for open~set action recognition.
\end{abstract}

%%%%%%%%% BODY TEXT
\section{Introduction}
\begin{figure}[t]
\centering
\includegraphics[width=0.85\linewidth]{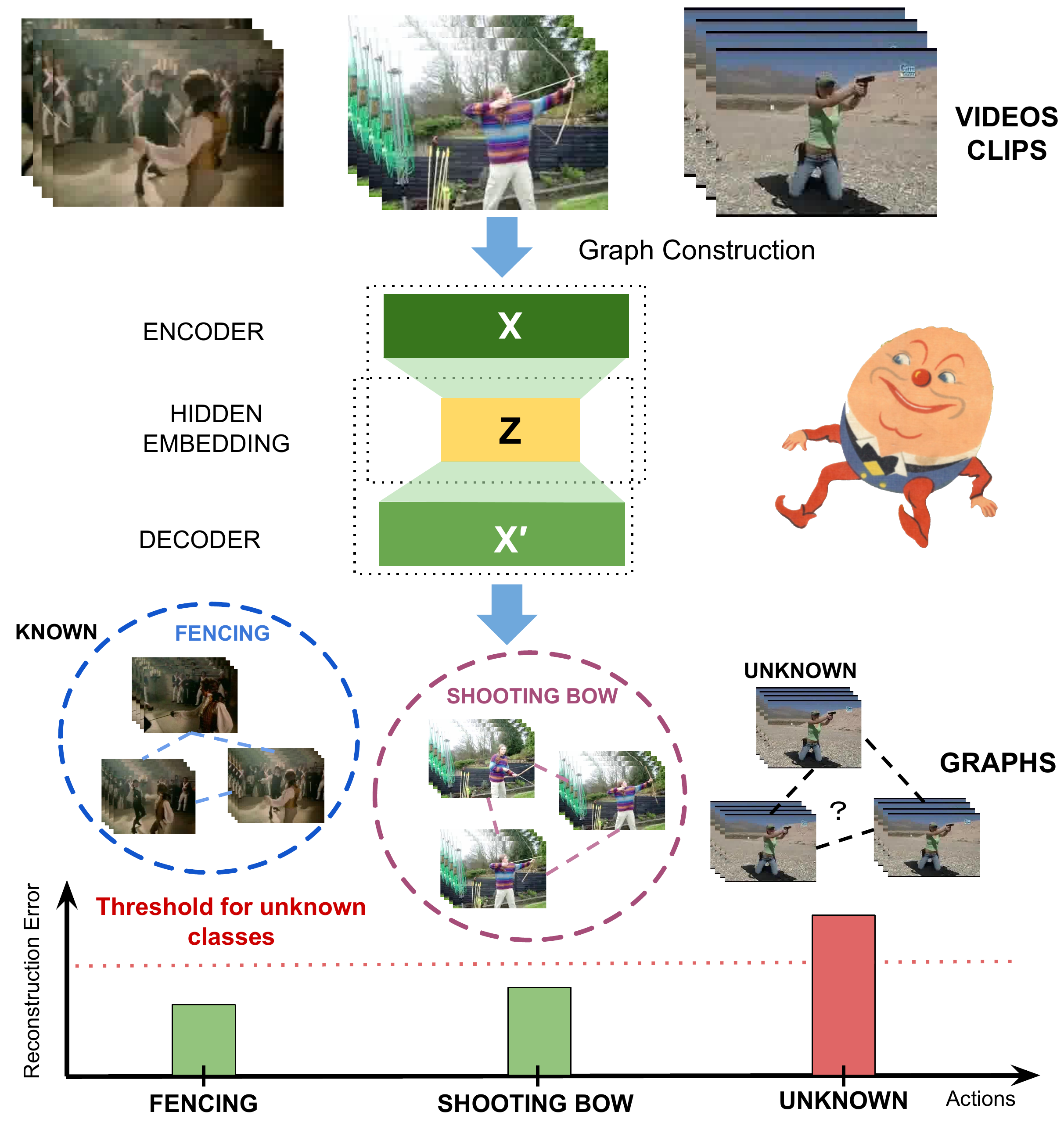}
\caption{Humpty Dumpty is trained to minimize the reconstruction error between the extracted relations $X$ in a video and the reconstructed relations $X'$. A large reconstruction error during inference is indicative of unknown relations, \ie, an unknown action, in the video. The namesake of our approach is the Humpty Dumpty nursery rhyme, which tells that when ``Humpty Dumpty had a great fall'' and broke, analogous to converting the clip into the latent feature space, ``all the King's horses and all the King's men couldn't put Humpty Dumpty'' back together again~\cite{carroll1897through} since his structure was unknown.}
\label{fig:concept}
\vspace{-10pt}
\end{figure}

Action recognition has garnered significant interest due to its potential applications in sports~\cite{soomro2014action}, surveillance~\cite{ji20123d}, smart homes~\cite{xia2012view}, etc. 
While significant progress has been made~\cite{simonyan2014two, wang2016temporal, DBLP:conf/iccv/TranBFTP15, feichtenhofer2016convolutional, DBLP:conf/cvpr/HaraKS18,DBLP:conf/cvpr/CarreiraZ17} in recognizing activities on action recognition datasets such as HMDB-51~\cite{DBLP:conf/iccv/KuehneJGPS11}, UCF-101~\cite{DBLP:journals/corr/abs-1212-0402}, and Kinetics-700~\cite{DBLP:conf/cvpr/CarreiraZ17}, almost all this progress has been made under a closed~world assumption~\cite{scheirer2012toward} such that the test data only contains instances of trained classes. 
In this closed~world setting, the algorithm predicts which of the known classes a test instance belongs to, and ignores the problem of whether the instance belongs to any of the unknown classes. 
The closed~world assumption is not realistic for many real-world problems, where knowing all classes is not feasible. 
Closed~world methods can be adapted to this open~set paradigm by predicting that a test instance is unknown when the maximal prediction of any known class is sufficiently weak. 
This is because, discriminatively-trained, closed~world models tend to be ineffective at defining boundaries against background data not included in training~\cite{jain2014multi}. 
This makes it difficult to tell the difference between an unusual instance of a known class and an instance of an unknown class.

To address these challenges, we develop an action recognition algorithm under the open~set premise~\cite{scheirer2012toward} that distinguishes known activities from unknown activities in the videos. 
Thereby, detecting novel activities~\cite{DBLP:conf/bmvc/GuneMBC19} in videos. 
Unknown activities are not present in the training set. While they might share some semantic or contextual characteristics with known activities, the relations among these characteristics are significantly different. 
Our approach models these characteristics, and their relations as nodes and edges of a graph. 
It learns to reconstruct the graph for known activities using a multi-feature graph autoencoder. 
During testing, we use the reconstruction error to determine if the activity occurring in the video is known or unknown.  

\begin{figure}[t]
\centering
\includegraphics[width=0.9\linewidth]{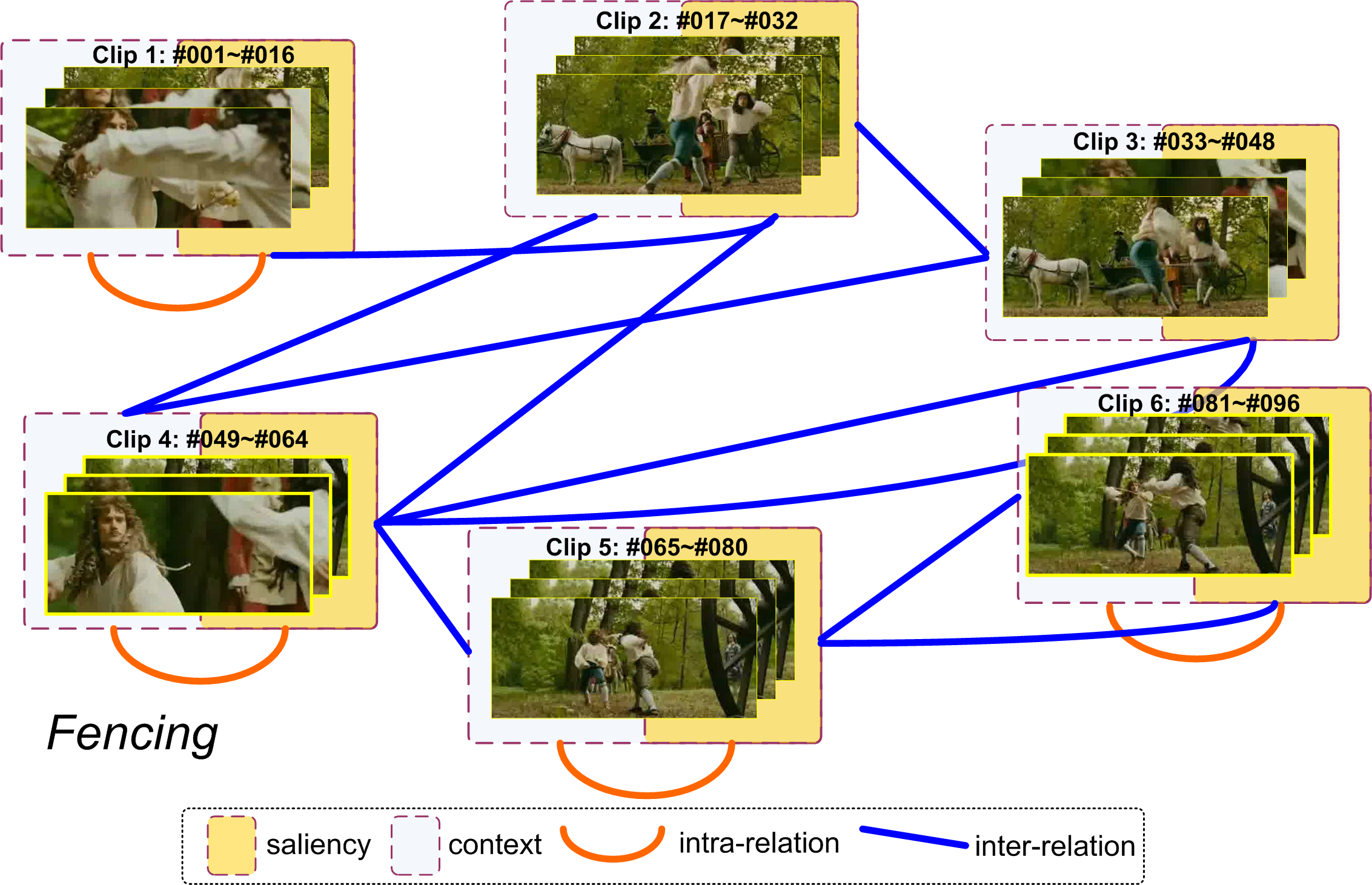}
\caption{The underlying semantic relations among clips in a \textit{Fencing} video. Video clips and the corresponding relations are partly displayed for more clarity.}
\label{fig:relation}
\vspace{-5pt}
\end{figure}

Recent work in open~set action recognition has relied on thresholding softmax scores~\cite{ramos2017detecting} or uncertainty in the classifier prediction~\cite{DBLP:conf/bmvc/RoitbergAS18} obtained from videos associated with known classes. 
A common theme among these techniques is treating the video as a single entity rather than exploiting the semantic relations within the video. 
Our approach captures these relations explicitly in a graph. To construct the graph, we divide the video into non-overlapping clips of fixed length. We obtain the contextual and semantic information for every clip using average pooling and max pooling, respectively, on the clip features. 
Using different pooling techniques to obtain multi-features or ``views'' for clustering is a common practice~\cite{DBLP:journals/isci/XueDDL19}. As shown in Figure~\ref{fig:relation}, each node in the graph denotes the pooled feature associated with a clip, while edges correspond to contextual or semantic relations between different clips or within the same clip. 
The graph is encoded in a latent space using a graph convolutional network (GCN), which effectively learns the common contextual and semantics relationships for known classes. The graph is then reconstructed during the decoding phase. We rely on the reconstruction error obtained from the graph autoencoder to estimate open~set risk directly based on these relations. 

We conduct extensive experiments on two benchmark datasets for action recognition: HMDB-51 \cite{DBLP:conf/iccv/KuehneJGPS11} and UCF-101 \cite{DBLP:journals/corr/abs-1212-0402}, where $50\%$ of the action labels are designated as novel for each dataset. 
We evaluate our method using the Receiver Operating Characteristic's Area Under the Curve (ROC-AUC) and the mean Average Precision (mAP). 
Humpty Dumpty achieves about $2\%$ and $3\%$ improvement in ROC-AUC scores along with $4\%$, and $1\%$ improvement in mAP scores over existing open~set action recognition methods \cite{DBLP:conf/nips/ScholkopfWSSP99, DBLP:journals/sigpro/PimentelCCT14,  DBLP:conf/iclr/HendrycksG17, DBLP:conf/bmvc/RoitbergAS18} on the HMDB-51 and UCF-101 datasets, respectively. 
Extensive ablation studies also show the impact of different components of our proposed method. 
The main contributions of our work are:
\begin{itemize}
\item A novel multi-feature graph to represent actions, constructed from recognizing similar contextual (average pooling) and salient (max pooling) relations across video clips for improved reconstruction;

\item A new method for open~set action recognition through open-space risk estimation~\cite{scheirer2012toward} that uses a graph-based autoencoder to learn the common semantics and features of known classes;

\item Extensive experiments and ablation studies showing superior performance over existing methods on two action recognition datasets that have been previously established as benchmarks for novel action recognition.
\end{itemize}

%-------------------------------------------------------------------------
\section{Related Work}
\subsection{Open~Set Recognition}
Open~set recognition~\cite{scheirer2012toward} is originally proposed in the image recognition domain and has received significant attention~\cite{winkens2020contrastive, wong2020identifying, perera2020generative} in recent years, including GMM \cite{DBLP:journals/sigpro/PimentelCCT14}, One-Class SVM~\cite{DBLP:conf/nips/ScholkopfWSSP99}, sparse coding~\cite{DBLP:journals/pami/ZhangP17}, extreme value theory~\cite{DBLP:conf/iccvw/MundtPMR19}, and CNNs~\cite{DBLP:conf/cvpr/PereraP19}. For example, Mundt~\etal~\cite{DBLP:conf/iccvw/MundtPMR19} combine model uncertainty with extreme-value theory for open~set recognition using prediction uncertainty on a variety of classification tasks. Perera~\etal~\cite{DBLP:conf/cvpr/PereraP19} learn globally negative filters from an external dataset and then threshold the maximal activation of the proposed network to identify novel objects effectively.

However, limited open~set work has been done in the action recognition domain. Recently, Roitberg~\etal~\cite{DBLP:conf/bmvc/RoitbergAS18} leverage the estimated uncertainty of individual classifiers in their predictions and propose a voting-based scheme to measure the novelty of a new input sample for action recognition. Shi~\etal~\cite{DBLP:conf/icmcs/0001WZYTS18} propose an open deep network to detect new categories by applying a multi-class triplet thresholding method, and then dynamically reconstructed the classification layer by continually adding predictors for new categories. 
Furthermore, Busto~\etal~\cite{DBLP:journals/pami/BustoIG20} propose an open~set domain adaptation method for action recognition where the target domain contains instances of categories that are not present in the source domain. These methods focus on modeling the novelty with respect to individual video clips within a video; which does not account for the semantic relations across clips. 
Our method, via its graph construction, explicitly learns how the spatial-temporal relations within a clip are temporally linked across the video clips, resulting in improved performance.

Although our approach focuses on open~set recognition, the relations captured by our approach can be used to detect anomalies in a video. 
It should be noted that there are fundamental differences between anomaly detection and open~set recognition. Anomalies are instance outliers, usually still within one of the known classes, that generally occur rarely, whereas unknowns/novelties come from different classes and can occur frequently during inference~\cite{DBLP:journals/corr/abs-1811-08581, DBLP:conf/bmvc/MasanaRSWL18}. The label spaces are also different - anomaly detectors produce frame level output while activity recognition algorithms label the entire video as an activity. 
Despite the difference in formulation, autoencoders~\cite{DBLP:conf/cvpr/0003CNRD16, DBLP:conf/mm/ZhaoDSLLH17, DBLP:conf/iccv/GongLLSMVH19} have been applied frequently for detecting anomalies in video. 

In addition, zero-shot learning focuses on the shared attributes between the known classes in training and the unknown classes during testing~\cite{DBLP:journals/corr/abs-1811-08581}, while the focus of open~set recognition is on determining the novelty of samples. They differ in the goal of the problem based on the information available during training and the desired result during testing. Therefore, a direct comparison would not be valid. There is a key difference in how each problem defines an unknown class - in zero-shot learning, unknown classes are specified through some shared side-information (such as attributes), while in open~set recognition the unknown classes are truly not specified in any way during training~\cite{DBLP:journals/corr/abs-1811-08581}. 

\begin{figure*}[t]
\centering
\includegraphics[width=0.95\linewidth]{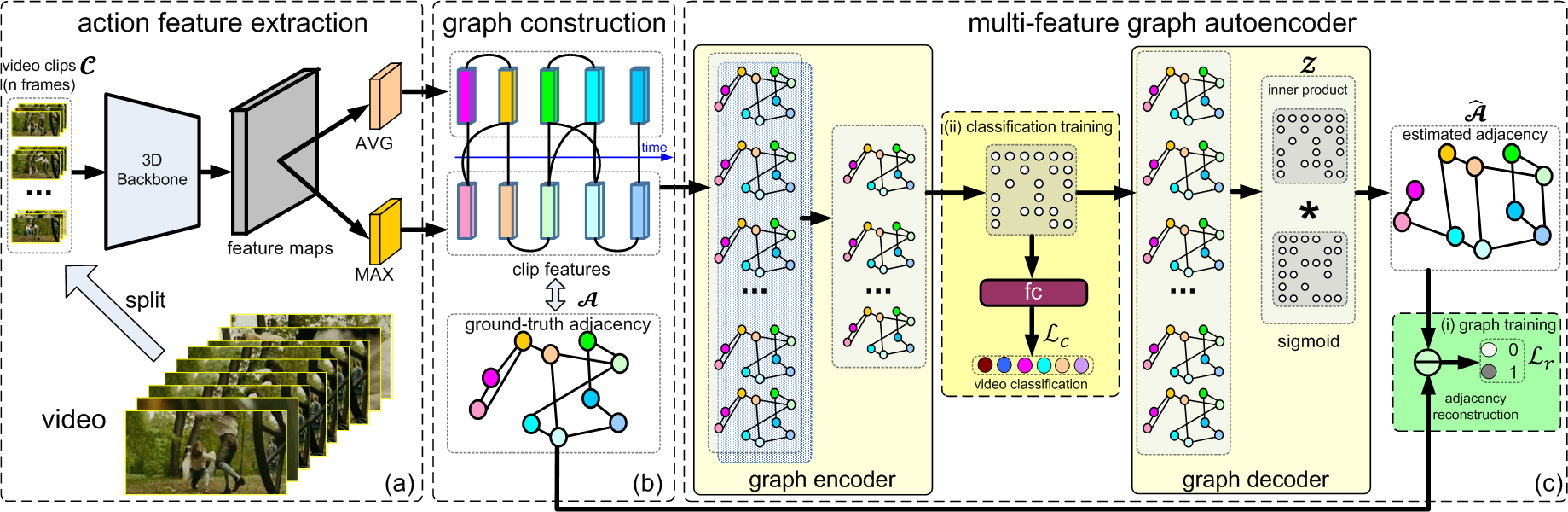}
\caption{Architecture of the Humpty Dumpty approach, including the (a) action feature extraction, (b) video graph construction, and (c) multi-feature graph autoencoder, as well as the two losses (i) reconstruction and (ii) known-class classification. The reconstruction error is used during evaluation to determine how the novelty of a given video. AVG and MAX stand for average pooling and max pooling layers respectively.}
\label{fig:method}
\vspace{-5pt}
\end{figure*}

\subsection{Graph Convolution Networks}
Increasing research attention has been devoted to generalizing neural networks~\cite{DBLP:journals/corr/BrunaZSL13,DBLP:journals/corr/KipfW16a,DBLP:journals/corr/abs-1905-10715,DBLP:conf/icml/GilmerSRVD17,DBLP:journals/tog/WangSLSBS19,DBLP:conf/iccv/Li0TG19} to work on arbitrarily structured graphs. The early prominent research~\cite{DBLP:journals/corr/BrunaZSL13} developed a graph convolution based on spectral graph theory. Kipf and Welling~\cite{DBLP:journals/corr/KipfW16a} developed a framework for unsupervised learning on graph-structured data based on the variational auto-encoder. In the work of Salehi and Davulcu~\cite{DBLP:journals/corr/abs-1905-10715}, the masked self-attentional layers are leveraged to attend over their neighbourhoods' features for a more discriminative representation. Gilmer~\etal~\cite{DBLP:conf/icml/GilmerSRVD17} propose message passing neural networks to exploit discriminative features from graph molecules. Wang~\etal~\cite{DBLP:journals/tog/WangSLSBS19} apply graph networks to point clouds by developing the EdgeConv layer. Recently, Li~\etal~\cite{DBLP:conf/iccv/Li0TG19} adapted residual/dense connections, and dilated convolutions to train deeper graph convolutional networks. Inspired by \cite{DBLP:journals/corr/abs-1905-10715}, we introduce a self-attention encoding scheme in the graph autoencoder to enhance the multi-features by weighting the nodes based-on information from neighboring nodes, achieving better graph reconstruction ability.

\subsection{Action Recognition}
In recent years, deep learning based methods~\cite{DBLP:conf/iccv/TranBFTP15, DBLP:conf/cvpr/HaraKS18, DBLP:conf/cvpr/CarreiraZ17} have dominated the field of video action recognition compared with traditional methods~\cite{DBLP:conf/iccv/LaptevL03, DBLP:conf/cvpr/WangKSL11, DBLP:conf/cvpr/SadanandC12}. Among these approaches, 3D CNNs~\cite{DBLP:conf/cvpr/HaraKS18,DBLP:conf/cvpr/TranWTRLP18} have been effective in encoding the temporal information in the video.
Tran~\etal~\cite{DBLP:conf/iccv/TranBFTP15} proposed temporal feature learning using deep 3D convolutional networks (C3D). 
3D ResNets~\cite{DBLP:conf/cvpr/HaraKS18} extended 2D ResNets~\cite{he2016deep} blocks and obtained better performance than the shallower C3D network. 
I3D~\cite{DBLP:conf/cvpr/CarreiraZ17} is a two-stream inflated 3D Convolutional Neural Network that uses RGB and Optical Flow to considerably improve the state-of-the-art in action recognition.
Feichtenhofer~\etal~\cite{DBLP:conf/iccv/Feichtenhofer0M19} proposed the SlowFast network for video recognition, including a Slow pathway to capture spatial semantics, and a Fast pathway to capture motion at fine temporal resolution. 
Moreover, Feichtenhofer developed another X3D~\cite{DBLP:conf/cvpr/Feichtenhofer20} backbone, which is progressively expanded from a tiny spatial ResNet network.

Since 3D CNNs model the temporal extent of a video using convolution operations, they have a fairly limited temporal range due to memory constraints on a GPU. Xi~\etal~\cite{ouyang20193d} combined 3D CNN with LSTM for action recognition. During graph construction, we use appearance similarity and temporal threshold to create an edge among the video clips. Wang~\etal~\cite{wang2018non} used a non-local self-attention layer to calculate temporal similarity for action recognition.
Our approach extracts temporal features from each non-overlapping video clip associated with the video. Unlike other approaches, we also then define contextual and semantic relations across these clips. 
%-------------------------------------------------------------------------
\section{Approach}
As discussed above, our task is open~set action recognition. Given a set of labeled videos with $\mathcal{L}_\mathcal{K}$ known classes available for training, and a set of unlabeled videos that contains a mixture of $\mathcal{L}_\mathcal{K}$ known classes along with $\mathcal{L}_{\mathcal{U}}$ unknown classes that appear only during testing, our goal is to classify the unlabeled video $x_j$ as either ``known'' if $x_j \in \mathcal{L}_\mathcal{K}$ or ``unknown'' if $x_j \in \mathcal{L}_{\mathcal{U}}$ by exploiting underlying semantic relations obtained from the training videos. To this end, our network represents intra-clip and inter-clip semantic relations associated with a video, and reconstructs the topographical information present in the graph. As illustrated in Figure~\ref{fig:method}, our model consists of three components: action feature extraction, video graph construction, and multi-feature graph autoencoder.

\subsection{Overview}
\label{subsec:overview}
As shown in Figure \hyperref[fig:method]{\ref*{fig:method}(a)}, we split the video into $\mathcal{C}$ clips such that the clips have no temporal overlap and each clip has $n$ image frames. For each clip $c \in \mathcal{C}$, we extract the feature vectors associated with all the $n$ frames in $c$. To obtain the multi-feature representation, we employ two kinds of pooling layers: maximal pooling and average pooling. The maximal pooling layer encodes the salient features present in the video clip, while the average pooling layer encodes the context information for the video clip. The multi-features are used to build the video graph, where each node represents the pooled feature obtained from a video clip and each edge connects two nodes with sufficient temporal and appearance similarity.

During the learning stage, a graph autoencoder is trained to reconstruct the graph obtained in the previous step. As with typical autoencoder methods, each training video provides its own groundtruth in the form of its graph. During the encoding phase, we apply a graph convolutional network (GCN) to map the video graph to a latent space. Furthermore, we employ self attention~\cite{DBLP:journals/corr/abs-1710-10903} to capture the most discriminative relations present in the graph. To learn to distinguish between known classes, we add a fully connected layer upon the latent space to classify the known instances in $\mathcal{L}_\mathcal{K}$ classes. In the decoding phase, the adjacency matrix is reconstructed by applying GCN over the latent space followed by a Sigmoid layer. During training, we minimize the reconstruction error between the original graph and the reconstructed graph. During inference (testing), the reconstruction error is used to measure the novelty of a video.

\subsection{Multi-Feature Graph Autoencoder}
\label{subsec:graph-construction}
The graph autoencoder is the crucial component of our approach. Inspired from the recent successes of relational modeling using a graph autoencoder~\cite{DBLP:journals/corr/KipfW16a}, we use it to reconstruct temporal clip similarities based on the underlying relations. 

\paragraph{Video Graph Construction.}
Given the multi-feature representation of all the clips in a video, we construct an undirected unweighted graph $\mathcal{G} =(\mathcal{V},\mathcal{E})$ to represent pair-wise relations among the video clips. Using the 3D backbone $B$, the pooled features obtained from a video clip are represented by the nodes and the semantic relations present between these features are represented by the edges, as shown in Figure~\hyperref[fig:method]{\ref*{fig:method}(b)}. Formally, the nodes are a set of pooled features, \ie, $\mathcal{V} = \{v_i \;| \; \forall v_i = \delta_\text{avg}(B(c_i))\cup\delta_\text{max}(B(c_i)), c_i \in \mathcal{C}\}$ where $\delta_\text{max}$, and $\delta_\text{avg}$ denotes the maximal and average pooling operations respectively. $\mathcal{C}$ is defined as the number of pooled node features within a video. The edge $e_{i,j} \in \mathcal{E}$ between nodes $v_i$ and $v_j$ is obtained using appearance similarity $f_a(v_i, v_j)$ and temporal distance $f_t(v_i, v_j)$ based on 
\begin{equation}
e_{i,j} = \mathbb{1}(f_a(v_i, v_j) \leq \theta_a, f_t(v_i, v_j) \leq \theta_t),
\label{equ:edge}
\end{equation}
where $\mathbb{1}(\cdot)=1$ if the argument is true and $0$ otherwise, and $\theta_a$ and $\theta_t$ are the thresholds for the appearance and temporal distance respectively. Thus the video graph can model the short-term temporal consistency and semantic similarity.
We use a binary adjacency matrix $\mathcal{A}$ to represent the edges of the graph when graph convolution is applied on the graph $\mathcal{G}$. The value $1$/$0$ at row $i$ and column $j$ of $\mathcal{A}$ indicates that there is an/no edge between node $v_i$ and $v_j$. The ground-truth adjacency matrix $\mathcal{A}$ is calculated as
\begin{equation}
\label{equ:adjacency}
\mathcal{A}(i,j)=\left\{
\begin{aligned}
&1 \quad \text{if} \quad e_{i,j}=1, \\
&0 \quad \text{otherwise}. \\
\end{aligned}
\right.
\end{equation}
In our experiments, we use the euclidean distance between the features associated with node $v_i$ and $v_j$ to calculate the appearance similarity $f_a(v_i, v_j)$. The temporal distance $f_t(v_i, v_j)$ is the absolute difference (counted by the number of frames) between the center frames for clips $c_i$ and $c_j$.  The node features $f$ are aggregated in a $C\times D$ matrix $\mathcal{F}$, where $D$ is the dimension of the clip feature. Since the graph is undirected, the adjacency matrix is symmetric, \ie, $\mathcal{A}(i,j)=\mathcal{A}(j,i)$.

\paragraph{Self-attention Graph Encoder.}
To learn the underlying semantic relations captured by the graph $\mathcal{G}$, we use a multi-feature graph autoencoder. 
As shown in Figure \hyperref[fig:method]{\ref*{fig:method}(c)}, the graph encoder learns a layer-wise transformation using the graph convolution (GC) over the the adjacency matrix $\mathcal{A}$ and the feature matrix $\mathcal{F}$ as follows.
\begin{equation}
\label{equ:encoder}
\mathcal{Z}^{l+1} = \text{GC}(\mathcal{Z}^{l},\mathcal{A};\mathcal{W}^{l}),
\end{equation}
where $\mathcal{Z}^{l}$ and $\mathcal{Z}^{l+1}$ are the input and output of the graph convolution at layer $l$ and $\mathcal{Z}^{0}=\mathcal{F}\in\mathbb{R}^{C\times D}$, the original graph constructed as above. $\mathcal{W}^{l}$ is the filter parameters of the corresponding graph convolutional layer in the network. For clarity, the index of layer $l$ is omitted in the following.

To enhance the discriminative representation of graph data, we further introduce the graph attention layer \cite{DBLP:journals/corr/abs-1710-10903} to perform self-attention encoding. The input of the graph attention layer is a set of node features $\mathcal{Z}=\{z_1,z_2,\cdots,z_{C}\}$; the output is a new set of node features with different cardinality $\mathcal{Z}'=\{z'_1,z'_2,\cdots,z'_{C}\}$. 
We use $K$ attention heads without sharing weights to obtain a diverse set of representations associated with the node features $\mathcal{Z}$. 
The node features $z_i$ are computed by aggregating the features associated with node $v_i$ and its neighbouring nodes. The aggregated feature are averaged over the attention heads to generate the updated feature $z'_i$ of node $v_i$, \ie,
\begin{equation}
z'_i = \sigma_s\big(\frac{1}{K}\sum_{k=1}^{K}\sum_{j\in\mathcal{N}_{v_i}\cup{v_i}}{\alpha_{ij}^k}\mathcal{W}_k z_j\big),
\label{equ:attention}
\end{equation}
where $\sigma_s$ denotes a non-linear softmax activation function. $\mathcal{N}_{v_i}$ denotes the neighbouring nodes of node $v_i$, and $\mathcal{W}_k$ is the corresponding input linear transformation's weight matrix. $\alpha_{ij}^k$ denotes normalized attention coefficients computed by the $k$-th attention head, which are computed as
\begin{equation}
\alpha^k_{ij} = \frac{\exp\big(\sigma_{l}(\alpha^T(\mathcal{W}_k z_i \doubleplus \mathcal{W}_k z_j))\big)}{\sum_{j\in\mathcal{N}_{v_i}\cup{i}}{\exp\big(\sigma_{l}(\alpha^T(\mathcal{W}_k z_i\ \doubleplus \mathcal{W}_k z_j))\big)}},
\label{equ:alpha}
\end{equation}
where $\doubleplus$ is the concatenation operation, and $\sigma_{l}$ is the leaky ReLU activation function. The weight vector $\alpha^T=\gamma(\mathcal{W}_k z_i,\mathcal{W}_k z_j)$ measures the importance of node $v_j$'s features to node $v_i$, which is calculated by a single-layer feed-forward neural network $\gamma$. As shown in Figure~\ref{fig:method}, we only use the self-attention layer in the first graph convolutional layer of the graph encoder.

\paragraph{Graph Decoder.}
After encoding the graph, we obtain the hidden embedding $\mathcal{Z}$ for graph reconstruction. As shown in Figure~\hyperref[fig:method]{\ref*{fig:method}(c)}, we use another graph convolutional layer to refine $\mathcal{Z}$ with higher cardinality. Followed by the dropout operation, we use the graph decoder to learn the similarity of each row in the hidden embedding to obtain the output adjacency matrix. Similar to the work of Kipf~and Welling~\cite{DBLP:journals/corr/KipfW16a}, we use the inner product to calculate the cosine similarity between each feature in the hidden embedding since it is invariant to the magnitude of features. In other words, by applying the inner product on the hidden embedding $\mathcal{Z}$ and $\mathcal{Z}^T$, we can learn the similarity of each node in $\mathcal{Z}$ and generate a reconstructed adjacency matrix $\hat{\mathcal{A}}$ as
\begin{equation}
\begin{aligned}
\hat{\mathcal{A}} = \text{Sigmoid}(\mathcal{Z}\mathcal{Z}^T).
\end{aligned}
\label{equ:decoder}
\end{equation}

\subsection{Training and Inference}
\label{subsec:graph-training}
The network is trained by minimizing the discrepancy between the ground-truth adjacency matrix $\mathcal{A}$ and the reconstructed adjacency matrix $\hat{\mathcal{A}}$.

\paragraph{Loss Function.}
To train the proposed network, we use two loss terms - (i) classification loss $\mathcal{L}_c$ and (ii) reconstruction loss $\mathcal{L}_r$. The classification loss penalizes the network for misclassifying known class samples. The reconstruction loss penalizes the network for poor reconstruction of the original adjacency matrix. The overall loss function is defined as
\begin{equation}
\begin{aligned}
\mathcal{L} &= \mathcal{L}_c+\mathcal{L}_r \\
&= \frac{1}{N}\sum_{i=1}^{N}\ell_{c}(\mathcal{P}(x_i), y_{x_i}; \mathcal{Z}_{x_i})+\frac{1}{N}\sum_{i=1}^{N}\ell_{r}(\mathcal{A}_{x_i}, \hat{\mathcal{A}}_{x_i}),
\end{aligned}
\label{equ:loss}
\end{equation}
where $\mathcal{P}(x_i)$ and $y_{x_i}$ are the softmax probability and the ground-truth class label of sample $x_i$ respectively. $\mathcal{Z}_{x_i}$ is the corresponding latent representation. The function $\ell_c$ is cross-entropy for the classification loss and the function $\ell_r$ is binary cross-entropy for the reconstruction loss.
\begin{figure*}[t]
\centering
     \begin{subfigure}[b]{0.46\textwidth}
         \centering
         \includegraphics[width=\textwidth]{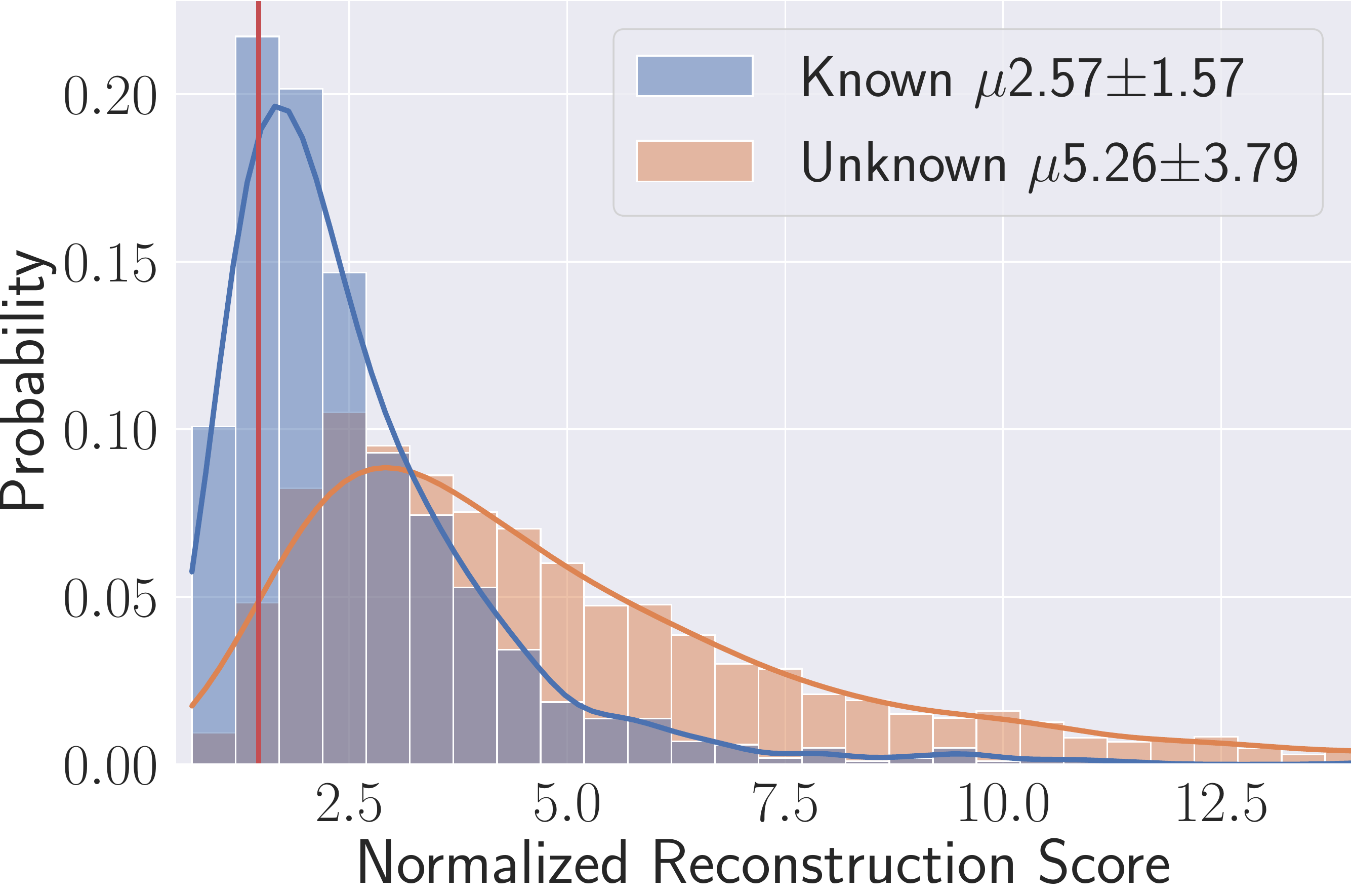}
         \caption{HMDB-51 dataset}
         \label{fig:reconstruction:hmdb}
     \end{subfigure}
     \begin{subfigure}[b]{0.46\textwidth}
         \centering
         \includegraphics[width=\textwidth]{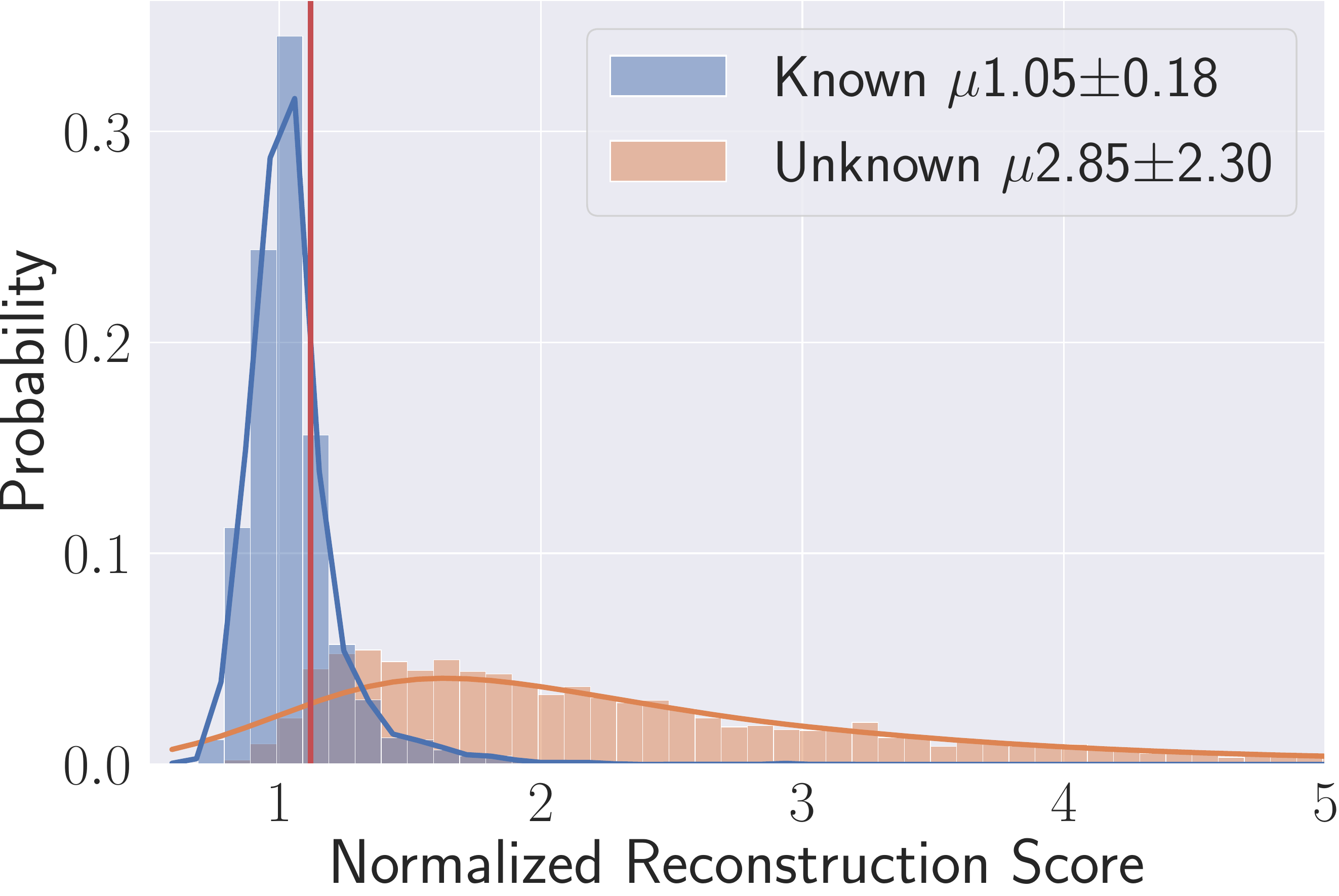}
         \caption{UCF-101 dataset}
         \label{fig:reconstruction:ucf}
     \end{subfigure}
\caption{Normalized Reconstruction Score (NRS) $\mathcal{S}(x_i)$ histograms for the known and unknown classes on the testing set. The x-axis indicates the normalized reconstruction score while the y-axis indicates the counts of instances. The \textcolor{BrickRed}{red} line denotes the max F1 score threshold for the dataset.}
\label{fig:reconstruction}
\vspace{-5pt}
\end{figure*}

\paragraph{Adjacency Matrix Re-construction.}
Since the length of every video varies in the dataset, we use the normalized reconstruction score $\mathcal{S}(x_i)$ of the graph $\mathcal{G}(x_i)$ to measure the novelty degree of the corresponding sample $x_i$, \ie,
\begin{equation}
\begin{aligned}
\mathcal{S}(x_i)=\beta\cdot\ell_r(\hat{\mathcal{A}}(x_i),\mathcal{A}(x_i)),
\end{aligned}
\label{equ:reconstruction}
\end{equation}
where $\hat{\mathcal{A}}$ and $\mathcal{A}$ are the estimated and ground-truth adjacency matrix respectively. $\ell_r$ is the binary cross-entropy function. To adjust to different sizes of adjacency, we use $\beta=\frac{N_\mathcal{V}^2-N_\mathcal{E}}{N_\mathcal{E}}$ as the normalization ratio. $N_\mathcal{V}$ and $N_\mathcal{E}$ denote the number of nodes and edges in graph $\mathcal{G}$ respectively.

\paragraph{Inference.}
To determine the open-space risk associated with a video, we compute the multi-feature representation of every video in the test set. These representations are passed through the graph autoencoder to determine the reconstruction error associated with the video. Given the ground-truth adjacency matrix computed by Equation \eqref{equ:adjacency}, we can estimate the reconstruction error to measure the degree of novelty by empirically setting the threshold. As shown in Figure~\ref{fig:reconstruction}, the video is regarded as ``unknown'' if the reconstruction error is above the threshold.

%-------------------------------------------------------------------------
\section{Experiments}
The proposed method is implemented using Pytorch~\cite{paszke2019pytorch} and all experiments are conducted on a workstation with an NVIDIA Titan X GPU card.

As discussed in the related work, few novelty detection methods have been used for video action recognition\footnote{Since there are no available source codes for the open~set action recognition methods \cite{DBLP:conf/icmcs/0001WZYTS18,DBLP:journals/pami/BustoIG20}, we cannot compare them in our experiment.}. 
According to \cite{DBLP:conf/bmvc/RoitbergAS18}, we compare our model to several baseline novelty detection methods. 
The One-Class SVM~\cite{DBLP:conf/nips/ScholkopfWSSP99} with an RBF kernel can model the videos of one (or all) known class(es), where the upper bound on the fraction of training errors $n$ is set to $0.1$. The Gaussian Mixture Model (GMM)~\cite{DBLP:journals/sigpro/PimentelCCT14} generatively represents the videos in sub-spaces with $8$ components, indicating novelty for any sample with sufficiently low probability. Softmax probabilities~\cite{DBLP:conf/iclr/HendrycksG17} of the I3D network~\cite{DBLP:conf/cvpr/CarreiraZ17} use the score output for a rejection thresholding to determine the novelty degree. Uncertainty \cite{DBLP:conf/bmvc/RoitbergAS18} is calculated by a Bayesian neural network (BNN) posterior distribution with different network parameters. Informed Democracy \cite{DBLP:conf/bmvc/RoitbergAS18} exploits the uncertainty of the output neurons in a voting scheme for novelty detection.

\begin{table*}[t]
  \centering
  \rowcolors{3}{gray!15}{white}
    \begin{tabular}{ccccc}
    \toprule
              Values reported as $\mu\pm\sigma$ & \multicolumn{2}{c}{HMDB-51~\cite{DBLP:conf/iccv/KuehneJGPS11}} & 
              \multicolumn{2}{c}{UCF-101~\cite{DBLP:journals/corr/abs-1212-0402}} \\
                %   \cmidrule(c){2-3} \cmidrule(c){4-5}
                   & ROC-AUC & mAP & ROC-AUC & mAP \\
    \midrule
    One-class SVM \cite{DBLP:conf/nips/ScholkopfWSSP99} & $54.09 (\pm3.0)$  & $77.86 (\pm4.0)$  & $53.55 (\pm2.0)$  & $78.57 (\pm2.4)$ \\
    Gaussian Mixture Model \cite{DBLP:journals/sigpro/PimentelCCT14}   & $56.83 (\pm4.2)$  & $78.40 (\pm3.6)$  & $59.21 (\pm4.2)$  & $79.50 (\pm2.2)$ \\
    Softmax Confidence \cite{DBLP:conf/iclr/HendrycksG17} & $67.58 (\pm3.3)$  & $84.21 (\pm3.0)$  & $84.28 (\pm1.9)$  & $93.92 (\pm0.7)$ \\
    Uncertainty \cite{DBLP:conf/bmvc/RoitbergAS18} & $71.78 (\pm1.8)$  & $86.81 (\pm2.5)$  & $91.43 (\pm2.3)$  & $96.72 (\pm1.0)$ \\
    Informed Democracy \cite{DBLP:conf/bmvc/RoitbergAS18} & $75.33 (\pm2.7)$  & $88.66 (\pm2.3)$  & $92.94 (\pm1.7)$  & $97.52 (\pm0.6)$ \\
    \midrule
    Humpty Dumpty (ours)  & $\bf 78.50 (\pm0.8)$ & $\bf 91.39 (\pm0.9)$ &  $\bf 94.84 (\pm1.2)$     & $\bf 98.38 (\pm0.3)$  \\
    \bottomrule
    \end{tabular}
    \caption{Open~set recognition results (mean and standard deviation over ten dataset splits).}
  \label{tab:results}%
%\vspace{-15pt}
\end{table*}%

\paragraph{Datasets.}
We use HMDB-51~\cite{DBLP:conf/iccv/KuehneJGPS11} and UCF-101~\cite{DBLP:journals/corr/abs-1212-0402} to evaluate our method. HMDB-51 is a video dataset with $6,766$ manually annotated videos and $51$ action classes. Each class includes at least $101$ clips, each with duration of at least $1$ second. It was collected from digitized movies and YouTube. UCF-101 is another large video dataset with $13,320$ clips and $101$ action classes. The clip length varies from $1.06$ second to $71.04$ second. The videos are collected from user-uploaded videos over the Internet. 
Following the work in \cite{DBLP:conf/bmvc/RoitbergAS18}, each dataset is split evenly into known/unknown classes. For a fair comparison, we use the same dataset splits as Roitberg~\etal~\cite{DBLP:conf/bmvc/RoitbergAS18} and formulate open~set action recognition as a binary classification problem. That is, we split HMDB-51 in $26/25$ known/unknown classes and UCF-101 in $51/50$ known/unknown classes.
Without determining the specific threshold, we use area under curve (AUC) values of the receiver operating characteristic (ROC), precision-recall (PR) curves and F1 scores to evaluate compared methods.

%-------------------------------------------------------------------------
\paragraph{Implementation Details.}
The evaluation datasets contain many short clips with durations of several seconds. The network is optimized using Adam with a learning rate of $1e-5$ for $300$ epochs. The similarity thresholds in Equation~\eqref{equ:adjacency} are determined empirically and set to $\theta_a=30$ and $\theta_t=2n=32$. The number of heads for self-attention encoding is set to $K=4$.

%-------------------------------------------------------------------------
\subsection{Result Analysis}
As presented in Table~\ref{tab:results}, it can be seen that our method outperforms existing approaches considerably. Our method achieves an ROC-AUC gain of $3.17\%$ and $1.90\%$ on the HMDB-51 and UCF-101 datasets, respectively. Meanwhile, the corresponding standard deviation is smaller than that in existing methods. This can be attributed to two reasons. First, we focus on modeling the temporal semantic relations among video clips instead of the appearance information of individual video clips. Second, the self-attention encoding scheme can capture the common representation among multi-features far more robustly.

\begin{figure*}[t]
\centering
\includegraphics[width=\linewidth]{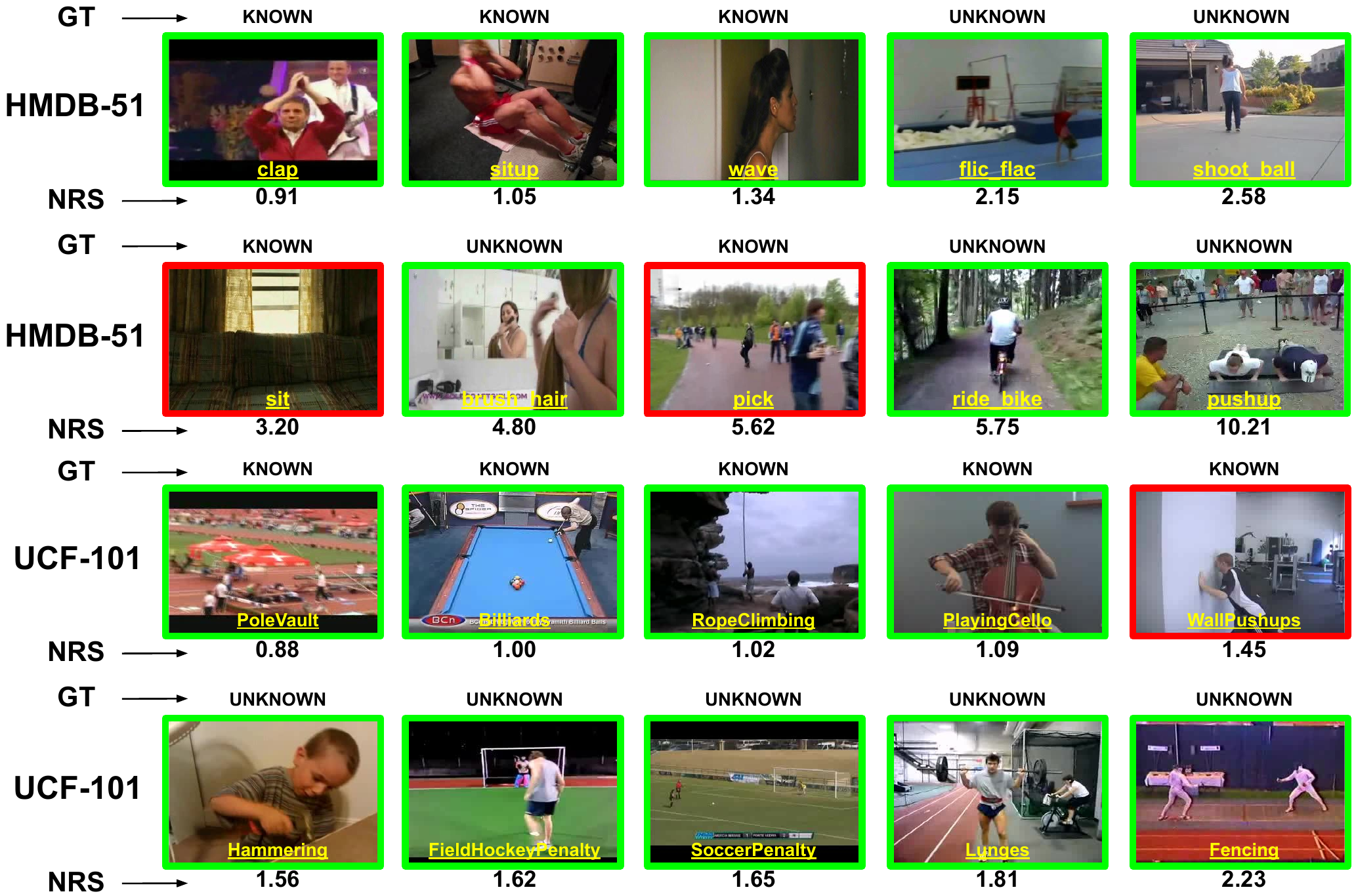}
\caption{Qualitative results on the HMDB-51 and UCF-101 datasets, where the videos are randomly selected. The color of the box denotes if our method correctly (\textcolor{Green}{green box}) or incorrectly (\textcolor{red}{red box}) identified whether the clip is from a known or unknown class.} 
\label{fig:qualitative}
\vspace{-5pt}
\end{figure*}

In Figure~\ref{fig:reconstruction}, we show the reconstruction score distributions for known and unknown classes on the testing sets of the above two datasets. The mean reconstruction scores of unknown classes are larger than that of known classes, $5.26$ vs. $2.57$ (Figure~\ref{fig:reconstruction:hmdb}). However, there exists some overlap between known and unknown distributions, resulting in weaker performance on HMDB-51. In contrast, the majority of reconstruction scores of known testing videos in UCF-101 is less than $1.00$ (Figure~\ref{fig:reconstruction:ucf}). This could be because we have fewer training samples on HMDB-51 than UCF-101. Figure \ref{fig:qualitative} shows some qualitative results obtained at different values of reconstruction error, where we chose the threshold exclusively for the qualitative visualization.

%-------------------------------------------------------------------------
\subsection{Ablation Studies} \label{subsec:ablation}
We further study the influence of three important components of our method: (a) multi-feature representation, (b) self-attention encoding, and (c) graph autoencoder. The ablation studies are conducted on a single split of HMDB-51.

\paragraph{Effectiveness of multiple features.}
To investigate the importance of the multiple feature representation, we test the system using only the average or the maximal pooling layer of the backbone (denoted as ``AVG'' and ``MAX'' in Figure~\hyperref[fig:method]{\ref*{fig:method}(a)}) to calculate the features of clips. From Table~\ref{tab:multiview}, the ``MAX'' variant performs similarly compared to the ``AVG'' variant. Moreover, using both features performs best by a comfortable margin in terms of all three metrics, a $2.1$ gain in ROC-AUC, $1.1$ gain in mAP and $0.67$ gain in max F1. We conclude that the multiple features are partly complementary in capturing semantic information in videos.
\begin{table}[t]
\centering
\begin{adjustbox}{max width=0.45\textwidth}
 \rowcolors{2}{gray!15}{white}
\begin{tabular}{cccc}
\toprule
Pooling Features & F1 Max      & ROC-AUC        & mAP      \\ \midrule
MAX    & 87.81          & 76.89          & 91.07          \\
AVG    & 87.39          & 76.61          & 91.06          \\
Multi-Feature  & \textbf{88.06} & \textbf{79.01} & \textbf{92.16} \\ \bottomrule
\end{tabular}
\end{adjustbox}
\caption{Open~set recognition results in terms of multiple features.}
\label{tab:multiview}
\vspace{-5pt}
\end{table}

\paragraph{Effectiveness of self-attention encoding.}
To verify the effectiveness of self-attention encoding, we enumerate the number of heads of self-attention encoding $K=\{1,2,4,6\}$ in Equation~\eqref{equ:attention}. $K=1$ implies that our method uses the convolutional graph layer instead of self-attention encoding to capture the relations among video clips. From the results shown in Table \ref{tab:attention}, the ROC-AUC and AP scores increase with $\mathcal{K}$ before the optimal performance ($K=4$) is achieved. This result indicates that self-attention encoding can model the temporal semantic relations among video clips effectively. However, the effect is not large and it is difficult for more heads of self-attention encoding to improve the performance. For $K=6$, we achieve slightly inferior ROC-AUC and mAP scores even though our F1 score improves. Due to better performance on both ROC-AUC and mAP scores, we set $K=4$ in our primary results.
\begin{table}[t]
\centering
\setlength{\tabcolsep}{15pt}
 \rowcolors{2}{gray!15}{white}
\begin{tabular}{cccc}
\toprule
\#$K$ & F1 Max & ROC-AUC & mAP     \\ \midrule
1                & 87.88                         & 76.38                        & 91.04          \\
2                & 88.07                         & 77.76                        & 91.38          \\
4                & 88.06                         & \textbf{79.01}               & \textbf{92.16} \\
6                & \textbf{88.14}                & 78.93                        & 92.04          \\ \bottomrule
\end{tabular}
\caption{Open~set recognition results in terms of different heads of self-attention encoding.}
\label{tab:attention} 
\vspace{-5pt}
\end{table}

\begin{table}[t]
\centering
\begin{adjustbox}{max width=\textwidth}
\rowcolors{2}{gray!15}{white}
\begin{tabular}{cccc}
\toprule
%\begin{tabular}[c]{@{}c@{}}Open~Set\\ Algorithms\end{tabular} & F1       & ROC-AUC        & mAP      \\ \hline
Open~Set Algorithms & F1 Max       & ROC-AUC        & mAP      \\ \midrule
EVM~\cite{DBLP:journals/pami/RuddJSB18}              & 87.49          & 69.24          & 86.51          \\
Auto-encoder~\cite{DBLP:conf/mm/ZhaoDSLLH17}      & 86.93          & 44.39          & 73.27          \\
Humpty Dumpty (ours)    & \textbf{88.06} & \textbf{79.01} & \textbf{92.16} \\ \bottomrule
\end{tabular}
\end{adjustbox}
\caption{Open~set recognition results of EVM, Autoencoder, and our method.}
\label{tab:gae}
\vspace{-5pt}
\end{table}

\paragraph{Effectiveness of graph autoencoder.}
To investigate the effectiveness of the graph autoencoder, we compare our method to two well-known open~set algorithms, the Extreme Value Machine (EVM)~\cite{DBLP:journals/pami/RuddJSB18} and Auto-encoder~\cite{DBLP:conf/mm/ZhaoDSLLH17}, which are trained on the same multiple features as that in Humpty Dumpty. As shown in Table~\ref{tab:gae}, we find a sharp decline in two metrics using EVM, \ie, $69.24$ vs. $79.01$ in ROC-AUC score, $86.51$ vs. $92.16$ in mAP score. This indicates that the hidden embedding in our graph autoencoder along with reconstruction error is more effective than the Weibull distribution used by EVM~\cite{DBLP:journals/pami/RuddJSB18}. 

Additionally, although our graph autoencoder and the standard autoencoder share similar concepts such as an encoder, decoder, and reconstruction, the standard autoencoder learns an approximation to the identity function such that the output features are similar to the original features, but it fails to capture the underlying relations among different features. In contrast, our graph autoencoder exploits contextual and semantic relations among video clips and then reconstructs the adjacency matrix of the video graph, resulting in much better performance. 

%------------------------------------------------------------------------
\section{Conclusion}
In this work, we propose a new multi-feature graph autoencoder, Humpty Dumpty, to address open~set action recognition. These results show that multiple features help the autoencoder learn a more robust reconstruction for known classes by exploiting salient and context information from the video. A self-attention step helps focus the network on the most important nodes, though too many self-attention heads have diminishing returns. Finally, we show that Humpty Dumpty can achieve state-of-the-art performance for open~set action recognition on two datasets.

{\noindent \textbf{Acknowledgements.}}
This material is based upon work supported by the Defense Advanced Research Projects Agency (DARPA) under Contract No. HR001120C0055. Any opinions, findings and conclusions or recommendations expressed in this material are those of the author(s) and do not necessarily reflect the views of the DARPA.

%%%%%%%%% REFERENCES
{\small
\bibliographystyle{ieee_fullname}
\bibliography{egbib}
}

\end{document}